\pdfoutput=1

\documentclass[11pt]{article}

\usepackage{EMNLP2023}

\usepackage{times}
\usepackage{latexsym}
\usepackage{booktabs}
\usepackage{graphicx}
\usepackage{colortbl}
\usepackage{array}
\newcommand{\diff}[1]{\colorbox{yellow!50}{#1}}
\usepackage{algorithm}
\usepackage{algpseudocode}
\usepackage{amsmath}
\usepackage{enumitem}
 \usepackage{soul}

\usepackage[T1]{fontenc}
\usepackage{amssymb}

\usepackage[utf8]{inputenc}

\usepackage{microtype}

\usepackage{inconsolata}

\title{Avoidance Decoding for Diverse Multi-Branch Story Generation}

\author{Kyeongman Park \\
Seoul National University \\
  \texttt{zzangmane@snu.ac.kr} \\\And
   Nakyeong Yang\\
  Seoul National University \\
  \texttt{yny0506@snu.ac.kr} \\\And
  Kyomin Jung\\
  Seoul National University \\
  \texttt{kjung@snu.ac.kr} \\}

\begin{document}
\maketitle

\begin{abstract}
Large Language Models (LLMs) often generate repetitive and monotonous outputs, especially in tasks like story generation, due to limited creative diversity when given the same input prompt.
To address this challenge, we propose a novel decoding strategy, \textbf{\textit{Avoidance Decoding}}, that modifies token logits by penalizing similarity to previously generated outputs, thereby encouraging more diverse multi-branch stories. This penalty adaptively balances two similarity measures: (1) Concept-level Similarity Penalty, which is prioritized in early stages to diversify initial story concepts, and (2) Narrative-level Similarity Penalty, which is increasingly emphasized later to ensure natural yet diverse plot development.
Notably, our method achieves up to $2.6$ times higher output diversity and reduces repetition by an average of 30\% compared to strong baselines, while effectively mitigating text degeneration.
Furthermore, we reveal that our method activates a broader range of neurons, demonstrating that it leverages the model's intrinsic creativity.
\end{abstract}

\section{Introduction}

Human writers can craft entirely different texts from the same ideas.
However, Large Language Models (LLMs) still struggle to reach human‑level creativity in writing.
Previous studies have found that even the state-of-the-art models such as GPT-4o \cite{openai2024gpt4o} generate repetitive and monotonous patterns \cite{zhang2025noveltybenchevaluatinglanguagemodels,wu2025depthneedexplorationiterative,wenger2025we,lagzian2025multi}. This tendency limits LLMs’ performance on tasks that require conceptual diversity and broad exploration, such as story generation \cite{park2024character,materzok2025cos,huang2024if} and complex reasoning \cite{wangtypedthinker,sun2025curiosity,kirk2023understanding,wu2025depth}. Especially in story generation, the task poses a unique challenge: it must engage readers through creative ideas and narratives, making it crucial to ensure diversity when generating stories with LLMs.

\begin{figure}[t] 
    \vspace{-0.4cm}
  \centering
  \includegraphics[width=0.9\columnwidth]{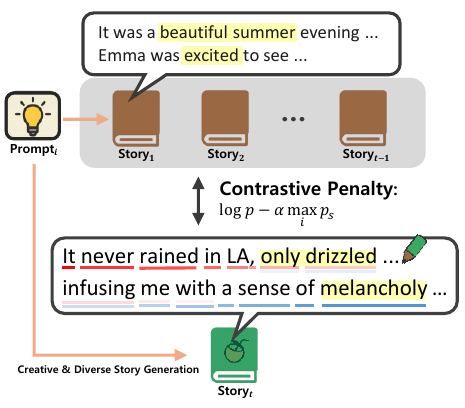}
  \caption{\textbf{\textit{Avoidance Decoding}} for discouraging similarity to previously generated stories.  \hl{Highlighted text} demonstrates the most contrast in the story induced by the Similarity-based Contrastive Penalty. The \textcolor{red}{red underlines} mark the front regions where the Conceptual Similarity Penalty is primarily applied, and the \textcolor{blue}{blue underlines} mark the backward regions where the Narrative Similarity Penalty is primarily applied.}
  \label{fig:intro}
\end{figure}

Existing studies have attempted to increase the diversity of generated texts through decoding-time methods \cite{welleck2019neural,holtzman2019curious,nguyen2024turning,vijayakumar2016diverse}. However, they have failed to achieve sufficient diversity since they only induce superficial token-level variations, without enriching conceptual, contextual, or narrative-level diversity. In addition, they have exhibited text degeneration due to an unresolved trade-off between diversity and fluency \cite{su2022contrastive,arias2024adaptive}.

To address these limitations, we propose a novel decoding strategy, \textbf{\textit{Avoidance Decoding}}.
Our method introduces a new Similarity-based Contrastive Penalty that modifies the model’s logits at each decoding step by penalizing similarity between the current output and multiple previously generated stories, which serve as negative samples.
As a result, our method can substantially increase the diversity of multi-branch stories, when given a single input prompt. Specifically, our Similarity-based Contrastive Penalty is a hybrid formulation of two distinct penalties: the Conceptual-level Similarity Penalty (\textbf{CSP}) and the Narrative-level Similarity Penalty (\textbf{NSP}). First, CSP is computed via similarity between the hidden states of candidate tokens for the next stepand negative samples to diversify initial concept representations. Second, NSP is computed via similarity between the sentence embeddings of the generated output and negative samples to ensure holistic narrative diversity. We assign higher weight to CSP in early stages to ensure diverse story planning, and gradually shift emphasis toward NSP as the current output length increases to ensure natural yet diverse plot progression, and apply the weighted sum as a final penalty.

As a result, without any additional training or stochastic sampling, our method achieves up to $2.6$ times higher diversity than the best baseline according to LLM‑based diversity evaluation. We also reduce repetition by at least 30\% on average compared to the baseline in automated metrics such as BLEU \cite{papineni2002bleu}, ROUGE‑L \cite{lin2004rouge}, METEOR \cite{banerjee2005meteor}, and sentence similarity \cite{reimers2019sentence}, while  maintaining robustness to text degeneration. Furthermore, we reveal that our method activates a broader range of neurons during iterative multi‑branch story generation. This indicates that it leverages the model’s intrinsic creative capacity, rather than merely introducing superficial token-level variations.


\section{Related Works}
\subsection{LLM Decoding Strategies}


Various decoding strategies exist for large language models (LLMs),  some focusing on promoting diversity \cite{welleck2019neural,holtzman2019curious, fan2018hierarchical, vijayakumar2016diverse, nguyen2024turning, zhu2023penalty}, increasing reliability \cite{hokamp2017lexically,wang2022self,chuang2023dola, guo2025dsvd,kim2025drift}, and achieving more fluent and high-quality generation \cite{meister2023locally,su2022contrastive,li2022contrastive,arias2024adaptive}. In our case, we utilize contrastive-style decoding strategy to enhance diversity during generation \cite{su2022contrastive,welleck2019neural,arias2024adaptive,li2022contrastive}; unlike prior work that reduces token-level similarity to previously generated tokens, we increase context-level diversity among multi-branch samples by penalizing similarity to each previously generated output.

\subsection{Diverse Story Generation}


There exist various methods to boost diversity and creativity in story generation \cite{park2024character,patel2024swag,fan2018hierarchical,bae2024collective,materzok2025cos, vijayakumar2016diverse}. Some approaches generate diverse story branches in a tree structure from a single prompt, enhancing creativity and engagement \cite{materzok2025cos,wen2023grove,nottingham2024improving,jaschek2019mysterious,alabdulkarim2021goal}. Others incorporate interactive story generation with human-in-the-loop branching at key decision points \cite{huang2024if,ghaffari2025narrative}. To the best of our knowledge, there is no method to enhance diversity during decoding time for multi-branch story generation, which is the main focus of our work.

\section{Problem Definition}
Given a fixed story prompt representing a core concept or initial idea of a story, \textbf{\textit{Multi-Branch Story Generation}} focuses on generating multiple coherent and fluent story continuations in parallel, which are mutually diverse in their narrative trajectories.
Suppose $P_{\theta}$ is a language model.
Given a fixed story prompt $p$, we denote $\{x^1, ..., x^n\}$ as the set of stories that are generated in parallel by $P_{\theta}$ conditioned on $p$.
Each $x^i$ is generated by predicting the next token $x^{i}_t$ from the distribution $P_{\theta}(x^{i}_{t} \mid p, x^{i}_{1:t-1})$, where $x^{i}_{1:t-1}$ denotes the previously generated tokens. 
We define the pairwise similarity function $s : \mathcal{X} \times \mathcal{X} \rightarrow \mathbb{R}$ where $\mathcal{X}$ denotes the space of all possible generated stories (e.g., token sequences), and $s(x_i, x_j)$ quantifies the semantic or structural similarity between two outputs $x_i$ and $x_j$, such as BLEU or Sentence-Similarity. The objective of this task is to generate a set of semantically divergent stories $\{x^1, ..., x^n\}$ from a single prompt $p$ by minimizing $s(x_i, x_j)$ for all $i \neq j$, thereby ensuring diversity in content and narrative structure while maintaining robustness against degeneration.

\section{Methodology}

We introduce a novel decoding strategy that steers the language model to generate outputs that do not resemble any of the negative samples while resisting degeneration. In this work, we treat previously generated outputs for the same input as negative samples, thereby enhancing diversity of multi-branch stories while mitigating degeneration.


\subsection{Motivation}
\label{motivaiton}

As a high-level justification for our method, we present a probabilistic motivation for contrastively modifying token logits using the maximum similarity penalty to negative samples. Suppose we have $P_{\theta}(x_{t}|x_{1:t-1}) = \prod_{j=1}^{t} P_{\theta}(x_{j}|x_{1:t-1})$ that is a language model, and $P_{s}(\mathcal{N}|x_{1:t})$ that represents the \textit{match probability} to quantify how closely the generated text (with the next candidate token appended) resembles the negative sample texts, $\mathcal{N} = \{n_{1}, ..., n_{N}\}$. We then define $P\bigl(x_t \mid x_{1:t-1}, \neg \mathcal{N}\bigr)$ as the probability of generating token \(x\) at step \(t\) given previous context \(x_{1:t-1}\) while explicitly avoiding any negative samples in $\mathcal{N}$, and
$P_{s}\bigl(\neg \mathcal{N} \mid x_{1:t}\bigr)$ as the complementary \textit{match-avoidance probability}, the probability that the generated text \(x_{1:t}\) does not match any of the negative samples \(\mathcal{N}\).
Ideally, this avoidance probability should take into account dependencies between negative samples: $P_s\bigl(\neg \mathcal{N} \mid x_{1:t}\bigr)
= \prod_{i=1}^N \bigl(1 - P_s(n_i \mid x_{1:t}, \neg n_{1:i-1})\bigr)$.
This formulation naturally reflects our main goal: generating outputs that are dissimilar to any previously generated stories (i.e., negative samples). However, in practice, we approximate each conditional match probability as $P_s(n_i \mid x_{1:t}, \neg n_{1:i-1}) \approx P_s(n_i \mid x_{1:t})$, to reduce computational cost. Substituting this into the product and applying Bayes’ rule yields:

\begin{equation}
\resizebox{\columnwidth}{!}{$
\begin{aligned}
P\bigl(x_{t}\mid x_{1:t-1},\neg \mathcal{N}\bigr)
&= \frac{P_{\theta}\bigl(x_t \mid x_{1:t-1}\bigr) P_{s}\bigl(\neg \mathcal{N} \mid x_{1:t-1}, x_t\bigr)}{P_{s}\bigl(\neg \mathcal{N} \mid x_{1:t-1}\bigr)} \\
&\propto P_{\theta}\bigl(x_t \mid x_{1:t-1}\bigr) P_{s}\bigl(\neg \mathcal{N} \mid x_{1:t}\bigr) \\
&=P_{\theta}\bigl(x_t \mid x_{1:t-1}\bigr)\prod_{i=1}^N\bigl(1 - P_{s}(n_{i}|x_{1:t})\bigr).
\end{aligned}
$}
\end{equation}

\noindent As $P_{s}\bigl(\neg \mathcal{N} \mid x_{1:t-1}\bigr)$ is constant when optimizing for $x_t$. We then have an approximated formula:

\begin{equation}
\resizebox{\columnwidth}{!}{$
\begin{aligned}
\log P\bigl(x_{t} \mid x_{1:t-1}, \neg \mathcal{N}\bigr)
  &\approx \log P_{\theta}(x_{t}\mid x_{1:t-1})
     \;-\;\sum_{i=1}^N \alpha_i\, P_{s}(n_{i}|x_{1:t}).
\end{aligned}
$}
\end{equation}

\noindent using a scaled first-order approximation \(\log(1 - p)\approx -\alpha_i\, p\) for each sample. Therefore, we apply a penalty to the original token logits $\ell_t$ (corresponding to the probability \(P(x_t \mid p, x_{<t})\)) resulting in the adjusted logits $\ell^*_t$ as follows:
\begin{equation}
  \ell^*_t
  = \ell_t \;-\;\sum_{i=1}^N \alpha_i\,P_{s}(n_{i}|x_{1:t}).
  \label{eq:multi-penalized-logits}
\end{equation}
The sum-based penalty in Eq. (\ref{eq:multi-penalized-logits}) considers all negative evidence, akin to an $L_1$-style regularization. To prevent over-penalization from negative-sample accumulation, we redefine the adjusted logits as $\ell^{\prime}_t$, akin to an $L_\infty$-style regularization:
\begin{equation}
  \ell^*_t = \ell_t \;-\;\max_{i=1}^N \alpha_i\,P_{s}(n_{i}|x_{1:t}). \label{eq:max-penalized-logits-new}
\end{equation}
Greedy decoding then proceeds using these modified logits as:
\begin{equation}
x^*_t= \operatorname*{argmax}_{v \in \mathcal{V}} \ell^*_t(v). \label{eq:multi-neg-greedy-modified}
\end{equation}

\subsection{Similarity-Based Contrastive Penalties}
The\textit{ match probability}, $P_{s}(\mathcal{N}|x_{1:t})$, corresponds to the degeneration penalty introduced in prior work, \textit{Contrastive Search} \cite{su2022contrastive}.
The prior work computes the degeneration penalty as the maximum cosine similarity between the hidden state of a candidate token $x_t$ and those of the preceding tokens $x_{1:t-1}$.
We extend this probability by aiming to degrade the similarity between the target output and multiple negative samples.
Furthermore, we also design novel penalty terms that account for both concept‑ and narrative‑level similarity to those samples.


\subsubsection{Concept-level Similarity Penalty}

We first propose the \textbf{CSP} (\textbf{C}oncept-level \textbf{S}imilarity \textbf{P}enalty) to diversify the concept representations of stories. Specifically, we calculate the CSP as maximum cosine similarity between last hidden representations of candidate tokens and those of all individual tokens of negative samples. Formally, we compute:

\begin{equation}
s_j^{\text{CSP}} = \max_{h' \in H^{-}} \text{cos}(h_j, h')
\end{equation}
where \(h_j\) is the last hidden state of the \(j\)-th candidate token, and \(H^{-}\) denotes the set of hidden states for all tokens in the negative samples.

This mainly encourages diversity in the low-level conceptual space, which significantly influences the diversity of the story planning.

\subsubsection{Narrative-level Similarity Penalty}
However, applying the CSP may degrade the overall coherence and naturalness of the story, as it directly disrupts the representations, particularly during the later stages of long-form generation. Therefore, we additionally propose the \textbf{NSP} (\textbf{N}arrative-level \textbf{S}imilarity \textbf{P}enalty) to encourage narrative distinction from negative samples.
The NSP is computed by the cosine similarity between the embedding of the current output sentence (after appending the candidate token) and each negative sample, using Sentence-Bert. Formally, we compute : 
\begin{equation}
s_{j}^{\mathrm{NSP}} = \cos\bigl(E(y_{1:t}\!\oplus\!w_j),\,E(x^{-})\bigr),
\end{equation}
where \(E(\cdot)\) denotes the Sentence-BERT embedding function, \(y_{1:t}\!\oplus\!w_j\) is the current output sentence after appending \(w_j\), and \(x^{-}\) is negative sample sentence.

This penalty can significantly enhance plot diversity by reducing semantic similarity at a higher-level context without compromising the story's coherence and naturalness.

\subsubsection{Concept-to-Narrative Hybrid Penalty}

However, in the early stages of generation, when the output has not yet formed a meaningful length of sentence, the NSP is often too small to be effective.

Therefore, we integrate Concept- and Narrative-level Similarity Penalty into a hybrid formulation. Specifically, we rely primarily on the CSP in the early stage, until the decoding step reaches the inflection point \(T_0\), ensuring diverse story planning. Then we progressively increase the weight of the NSP to ensure the coherence and naturalness of the generated stories. We define the mixing ratio \(\gamma\) as follows:

\begin{equation}
\gamma = \delta + (1-\delta) \cdot \mathrm{sigmoid}\bigl(t - T_0\bigr)
\label{gamma_cal}
\end{equation}

\noindent where $\delta$ denotes the minimum weight assigned to CSP. Then the Concept-to-Narrative Hybrid Penalty is computed as:

\begin{equation}
s_{j}^{\mathrm{hybrid}} = \gamma \, s_{j}^{\mathrm{CSP}} \;+\; (1 - \gamma)\, s_{j}^{\mathrm{NSP}}
\label{penalty_cal}
\end{equation}

\noindent The term $s_{j}^{\mathrm{hybrid}}$ corresponds to the match probability $P_{s}\bigl(n_{i}\mid x_{1:t}\bigr)$ in Equation~\ref{eq:max-penalized-logits-new}, after appending the candidate token $w_{j}$ to the generated text $x_{1:t-1}$.


\subsection{Overall Decoding Procedure}

As shown in Algorithm~\ref{algorithm}, we first compute the number of candidate tokens $k$ and the adaptive penalty weight $\alpha_{ACS}$, following the process in the prior study~\cite{arias2024adaptive}. Next, we apply Equations~\ref{gamma_cal} and~\ref{penalty_cal} to compute the penalty for each candidate token with respect to each negative sample, scaling it by a constant hyperparameter $\beta$, and take the maximum value across them. Finally, we compute the final score $F_j$ for each candidate by combining the logit probability $p_t(w_j)$ with the penalty term $s_j^{\text{final}}$, weighted by  $\alpha_{\text{ACS}}$, and select the token with the highest score via greedy decoding.

\begin{algorithm}[H]
\caption{Avoidance Decoding}
\renewcommand{\arraystretch}{0.9}
 \hspace*{\algorithmicindent} \textbf{Input :}  $p_t$, $y_{1:t}$, $\{x_i^{-}\}_{i=1}^N$, $\{H^{-}_i\}_{i=1}^N$
\begin{algorithmic}[1]

\State Compute $k$, $\alpha_{ACS}$  \cite{arias2024adaptive}
\State Select top-$k$ tokens $\{w_j\}_{j=1}^k$ from $p_t$
 \State $\gamma = \delta + (1-\delta) \cdot \mathrm{sigmoid}(t-T_0)$
\For{$j = 1$ to $k$}
        \State $h_j \gets$   last hidden state of $\text{model}(w_j \mid y_{1:t})$
        \For{$i = 1$ to $N$}
        \State $s_j^{\text{CSP}} [i] \leftarrow  ( \max_{h' \in H_i^{-}} \text{cos}(h_j, h') )$
        \State  $s_j^{\text{NSP}} [i] \gets  \cos(E(y_{1:t}\!\oplus\!w_j), E(x_i^{-}))$
       
        \State $s_{j}^{\mathrm{hybrid}}[i] \gets \gamma \cdot s_j^{\text{CSP}}[i]+ (1 - \gamma) \cdot s_j^{\text{NSP}}[i]$
        \EndFor
    \State $s_j^{\text{final}} \gets \max_{i} \beta \cdot s_{j}^{\mathrm{hybrid}}[i]$
    \State $F_j \gets (1 - \alpha_{ACS} ) \cdot p_t(w_j) - \alpha_{ACS} \cdot s_j^{\text{final}}$
\EndFor
\State $w^* \gets \arg\max_j F_j$
\State \Return $w^*$
\end{algorithmic}
\label{algorithm}
\end{algorithm}

\noindent where $H^{-}_i$ is the set of last hidden states of all individual tokens in i-th negative sample, $p_t$ is the model’s logits, $y_{1:t}$ is generated tokens so far, and $x_i^{-}$ is i-th negative sample text.

\section{Experiments}

\begin{table*}[t]
\centering
\renewcommand{\arraystretch}{0.85} 
\setlength{\tabcolsep}{12pt}  
\resizebox{\textwidth}{!}{
\begin{tabular}{lcccccc}
\toprule
\textbf{method} & \textbf{BLEU(↓)} & \textbf{RougeL(↓)} & \textbf{METEOR(↓)} & \textbf{Sent-Sim(↓)} & \textbf{LLMScore(↑)} & \textbf{Degen(↓)} \\
\midrule
Naive                & 2.21  & 16.19 & 20.95 & 48.51 & 20.75 & 0.02 \\
Top-k               & 1.92  & 15.53 & 21.09 & 50.47 & 22.60 & 0.01 \\
Top-p               & 1.25  & \textbf{10.80} & 14.68 & 47.22 & 23.25 & 0.09 \\
Typical             & 1.47  & 14.13 & 18.76 & 50.54 & 21.75 & 0.02 \\
Mirostat            & 9.64  & 27.94 & 31.02 & 55.16 & 19.75 & 0.01 \\
Min-p               & 12.30 & 28.96 & 33.26 & 53.97 & 19.25 & 0.00 \\
CS                  & 52.14 & 65.70 & 67.35 & 66.18 & 17.60 & 0.00 \\
ACS                 & 50.71 & 63.80 & 66.58 & 63.12 & 20.50 & 0.00 \\
DBS                 & 11.99 & 25.04 & 31.20 & 58.81 & 25.25 & 0.03 \\
\midrule
GPT‑4o              & 3.57  & 18.48 & 24.61 & 51.89 & 22.25 & 0.04 \\
\midrule
Ours$_{CSP}$  & 0.61  & 9.31  & 9.95  & 25.25 & 69.75 & \underline{0.15} \\
Ours$_{NSP}$    & 39.87 & 48.71 & 51.23 & 71.44 & 25.25 & 0.00 \\
\rowcolor{gray!20}
\textbf{Ours}       & \textbf{1.04} & 12.57 & \textbf{14.36} & \textbf{27.56} & \textbf{65.40} & 0.02 \\
\bottomrule
\end{tabular}}
\caption{ReedsyPrompts, Mistral 7B}
\label{main_rp_mistral7b}
\end{table*}

\begin{table*}[t]
\centering
\renewcommand{\arraystretch}{0.85} 
\setlength{\tabcolsep}{12pt}  
\resizebox{\textwidth}{!}{
\begin{tabular}{lcccccc}
\toprule
\textbf{method} & \textbf{BLEU(↓)} & \textbf{RougeL(↓)} & \textbf{METEOR(↓)} & \textbf{Sent-Sim(↓)} & \textbf{LLMScore(↑)} & \textbf{Degen(↓)} \\
\midrule
Naive                   & 3.32  & 14.69 & 20.12 & 56.22 & 30.75 & 0.03 \\
Top‑k                  & 2.36  & 16.85 & 22.58 & 57.30 & 21.85 & 0.01 \\
Top‑p                  & 7.22  & 23.83 & 29.50 & 59.34 & 19.50 & 0.00 \\
Typical                & 1.92  & 15.18 & 19.93 & 58.84 & 21.50 & 0.02 \\
Mirostat               & 15.64 & 34.05 & 37.27 & 65.25 & 19.60 & 0.02 \\
Min‑p                  & 15.29 & 33.55 & 36.70 & 64.13 & 26.85 & 0.00 \\
CS                     & 72.02 & 80.78 & 81.62 & 79.88 & 15.25 & 0.00 \\
ACS                    & 72.49 & 81.89 & 83.07 & 79.09 & 17.00 & 0.01 \\
DBS                    & 13.40 & 27.14 & 33.12 & 66.79 & 21.60 & 0.04 \\
\midrule
GPT‑4o                 & 3.57  & 18.48 & 24.61 & 51.89 & 24.15 & 0.00 \\
\midrule
Ours$_{CSP}$     & 0.80  & 10.83 & 11.62 & 29.51 & 51.50 & \underline{0.12} \\
Ours$_{NSP}$       & 35.12 & 45.66 & 48.75 & 73.31 & 23.50 & 0.00 \\
\rowcolor{gray!20}
\textbf{Ours}          & \textbf{1.45} & \textbf{13.96} & \textbf{15.72} & \textbf{35.42} & \textbf{44.90} & 0.05 \\
\bottomrule
\end{tabular}}
\caption{Writing Prompts, Mistral 7B}
\label{main_wp_mistral7b}
\end{table*}

\begin{table*}[t]
\centering
\renewcommand{\arraystretch}{0.8}
\setlength{\tabcolsep}{12pt}  
\resizebox{\textwidth}{!}{
\begin{tabular}{lcccccc}
\toprule
\textbf{method} & \textbf{BLEU(↓)} & \textbf{RougeL(↓)} & \textbf{METEOR(↓)} & \textbf{Sent-Sim(↓)} & \textbf{LLMScore(↑)} & \textbf{Degen(↓)} \\
\midrule
Top‑k               &  1.12 & \textbf{11.85} & 17.94 & 48.49 & 34.25 & 0.04 \\
Top‑p               &  6.22 & 17.96          & 24.34 & 52.50 & 29.35 & 0.00 \\
Min‑p               & 20.05 & 30.30          & 36.20 & 60.27 & 26.85 & 0.00 \\
ACS                 & 59.78 & 64.91          & 67.68 & 77.95 & 19.00 & 0.00 \\
\midrule
GPT‑4o              &  3.57 & 18.48          & 24.61 & 51.89 & 22.25 & 0.04 \\
\midrule
Ours$_{CSP}$  &  0.85 & 11.61          & 14.06 & 31.73 & 60.10 & \underline{0.23} \\
Ours$_{NSP}$    & 31.16 & 40.02          & 43.65 & 68.16 & 27.20 & 0.00 \\
\rowcolor{gray!20}
\textbf{Ours}       & \textbf{1.09} & 12.40   & \textbf{15.63} & \textbf{32.66} & \textbf{54.25} & 0.09 \\
\bottomrule
\end{tabular}}
\caption{ReedsyPrompts, Llama 3B}
\label{main_rp_llama3b}
\end{table*}

\begin{table*}[t]
\centering
\renewcommand{\arraystretch}{0.8}
\setlength{\tabcolsep}{12pt}  
\resizebox{\textwidth}{!}{
\begin{tabular}{lcccccc}
\toprule
\textbf{method} & \textbf{BLEU(↓)} & \textbf{RougeL(↓)} & \textbf{METEOR(↓)} & \textbf{Sent-Sim(↓)} & \textbf{LLMScore(↑)} & \textbf{Degen(↓)} \\
\midrule
Top‑k               &  1.12 & 12.18 & 18.53 & 51.74 & 32.00 & 0.05 \\
Top‑p               &  4.66 & 16.98 & 23.67 & 55.70 & 27.00 & 0.08 \\
Min‑p               & 42.59 & 48.20 & 53.44 & 77.84 & 17.90 & 0.00 \\
ACS                 & 68.96 & 72.09 & 73.45 & 86.12 & 18.50 & 0.01 \\
\midrule
GPT‑4o              &  3.57 & 19.71 & 26.11 & 58.97 & 24.15 & 0.00 \\
\midrule
Ours$_{CSP}$  &  0.82 & 11.54 & 14.03 & 32.96 & 57.25 & \underline{0.13} \\
Ours$_{NSP}$    & 28.60 & 38.13 & 42.54 & 69.82 & 23.25 & 0.00 \\
\rowcolor{gray!20}
\textbf{Ours}       & \textbf{1.05} & \textbf{11.68} & \textbf{15.10} & \textbf{34.03} & \textbf{50.60} & 0.09 \\
\bottomrule
\end{tabular}}
\caption{Writing Prompts, Llama 3B}
\label{main_wp_llama3b}
\end{table*}

\subsection{Experimental Setup}

\paragraph{Implementation Details}

We run every decoding process on two NVIDIA RTX A5000 GPUs. To accelerate iterative token generation, we reuse the cached KV values of the Attention modules from the previous decoding step for the next one. We collect 20 versatile story prompts from ReedsyPrompts \cite{park2024longstory} and WritingPrompts \cite{fan2018hierarchical} that yielded at least 20 different human stories, and use these prompts as our initial inputs. We set the constant scalar hyperparameter $\beta$ to 2.0 to meaningfully influence the token logits, and $\delta$ to 0.5 to maintain sufficient low-level diversity throughout the generation process, based on empirical tuning (See Appendix \ref{beta_delta_appendix}).

\paragraph{Baselines}
We compare our method against several strong baselines:\vspace{-0.2cm}
\begin{itemize}[leftmargin=0.14in]
    \item \textbf{Naive, Top-k, Top-p, Typical, Mirostat, and Min-P sampling \cite{holtzman2019curious,meister2023locally,basu2020mirostat, nguyen2024turning}:} These are various stochastic decoding strategies that can enhance diversity.  Note that for Naive sampling, we apply no special techniques other than adjusting the temperature.\vspace{-0.2cm}
    \item \textbf{Contrastive Search (CS) and Adaptive Contrastive Search (ACS):} We include the original Contrastive Search and its advanced version, Adaptive Contrastive Search, as baselines. These methods serve as the primary motivation for our proposed methodology.\vspace{-0.2cm}
    \item \textbf{Diverse Beam Search (DBS) \cite{vijayakumar2016diverse}: } For Diverse Beam Search, we set the number of beam groups equal to the number of candidate sentences to maximize diversity.\vspace{-0.2cm}
    \item \textbf{GPT-4o:} We include OpenAI's powerful language model, GPT-4o, as a strong baseline.\vspace{-0.2cm}
    \item \textbf{Ours$_{CSP}$ and Ours$_{NSP}$ :} To demonstrate the effectiveness of our Hybrid Penalty, we include two ablated versions as baselines. Ours$_{CSP}$ utilizes only CSP as the penalty, while Ours$_{NSP}$ utilizes only NSP as the penalty. Note that Ours$_{CSP}$ is highly inspired by the formulation of prior work (\textit{CS}), yet integrates our own adjustments.
\end{itemize}

For all baselines except the ablated versions (i.e., Ours$_{CSP}$ and Ours$_{NSP}$), we feed all previously generated outputs from the same story prompt back into the next input and instruct the model to ``create a story that does not resemble any of the already generated outputs.'' See Appendix \ref{appendix:instruction_baseline} for more detailed information. In contrast, all Ours variants receive the exact same instruction at each iteration step within a story prompt, while internally storing all generated outputs in a \textit{Negative Examples} memory and computing the maximum hidden‑state similarity and sentence‑similarity from them.

\begin{figure}[t]     
  \centering
  \includegraphics[width=\columnwidth]{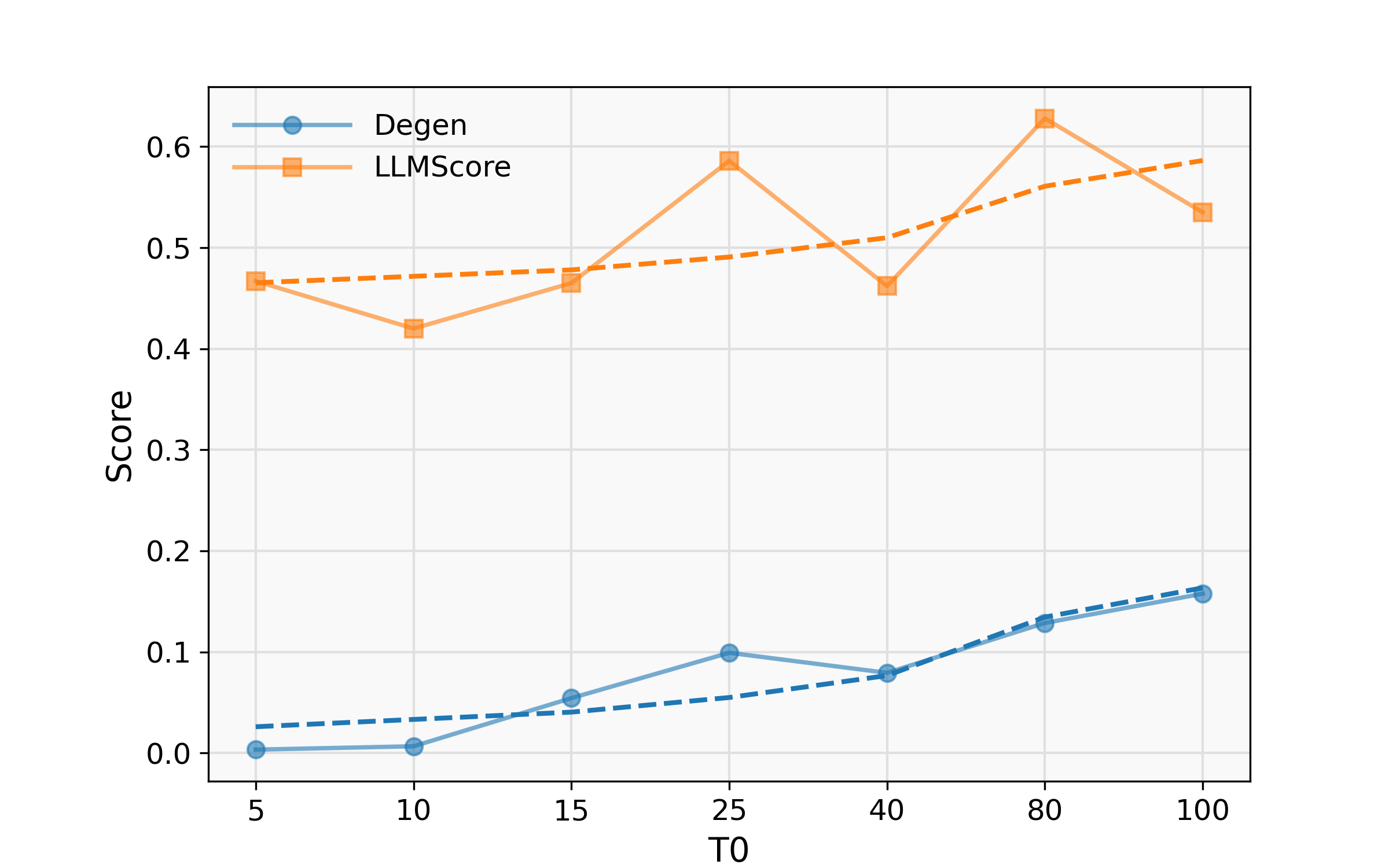}
  \caption{Average Degeneration and LLMScore versus $T_{0}$.}
  \label{fig:L0_test}
\end{figure}

\begin{figure}[t]     
  \centering
  \includegraphics[width=\columnwidth]{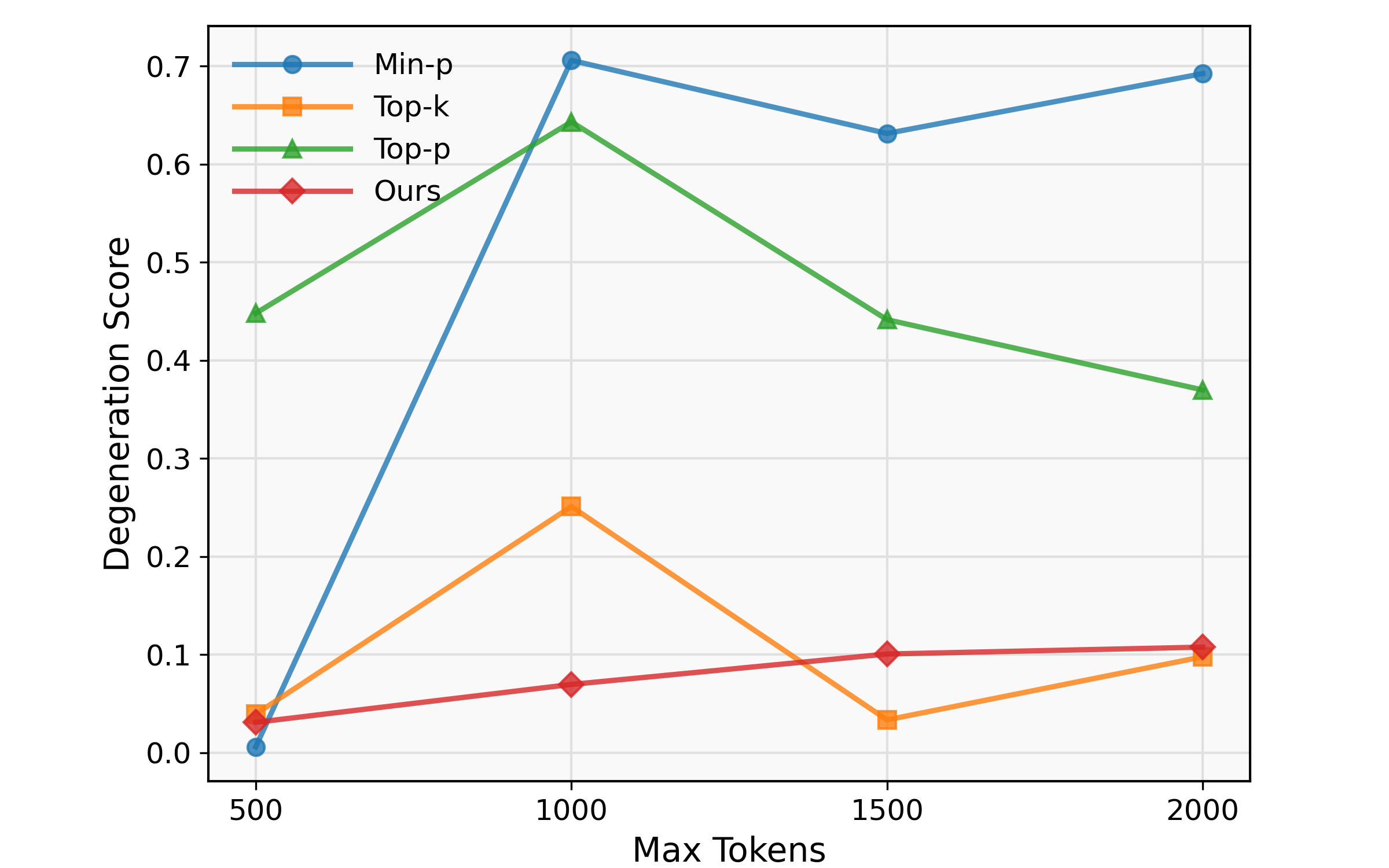}
  \caption{Average Degeneration scores versus max tokens.}
  \label{fig:max_token}
\end{figure}

\begin{figure}[t]     
  \centering
  \includegraphics[width=\columnwidth]{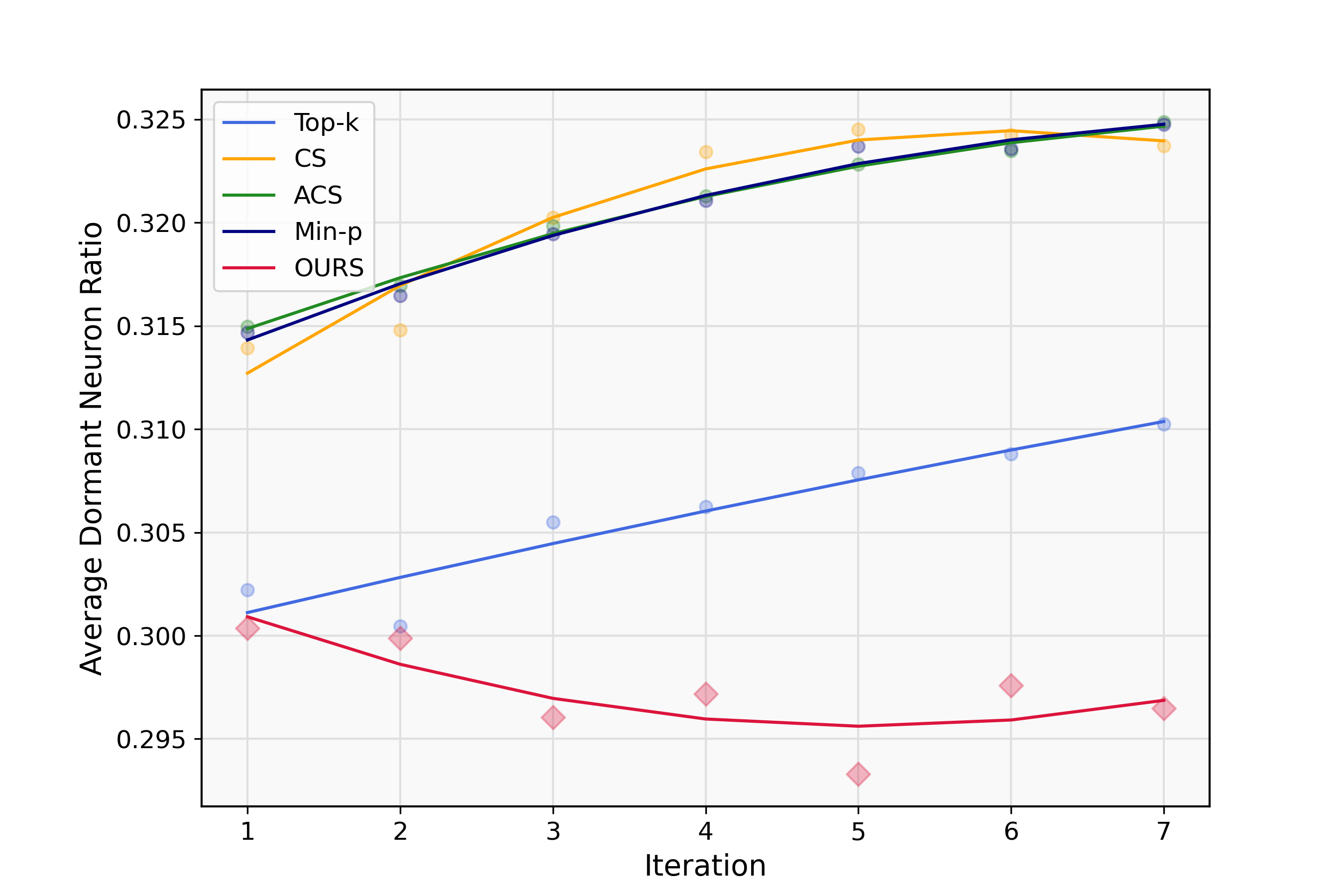}
  \caption{Average dormant neuron ratios per iteration.}
  \label{fig:dormant_ratio}
\end{figure}

\begin{table*}[t]
\centering

\begin{tabular}{>{\raggedright\arraybackslash}m{0.45\textwidth}  >{\raggedright\arraybackslash}m{0.45\textwidth}}
\hline
\multicolumn{2}{l}{\textbf{Prompt:} Write a story from the different perspectives of two people meeting for a blind date.} \\
\rowcolor{gray!20}
\textbf{Negative Sample} & \textbf{ Generation Result} \\
\hline
It was a \diff{beautiful summer evening}, and the sun was setting over the \diff{bustling city}. Emma was \diff{excited} to see if their chemistry was real, but she was also a \diff{little nervous}. (...) Meanwhile, Ryan was getting ready for the date, feeling a \diff{mix of excitement and nerves}. He had been looking forward to this all week, but he was also a little... &

On this particular evening, \diff{the drizzle} seemed to seep into the very marrow of my bones, infusing me with a sense of \diff{melancholy} (...) I stood outside the \diff{nondescript bistro}, clutching my glass of Pinot Grigio (...) ``So, how's your day been?'' ``What do you like to do in your free time?'' Ugh, who came up with these \diff{boring} conversation starters? ...\\
\hline
\end{tabular}
\caption{Two multi-branch stories generated from the same prompt. The left story is a prior output, and the right story is a new generation. We highlighted the most contrastive parts, including mood, setting, and emotions.}
\label{tab:qualitative_analysis}
\end{table*}

\subsection{Main Experiment Results}
\subsubsection{Automatic Evaluation Metrics}
We assess the performance of baseline methods using standard automatic evaluation metrics. Specifically, we first compute the average similarity among all stories generated for each prompt. The final score is obtained by averaging these values across all prompts. \vspace{-0.2cm}
\begin{itemize}[leftmargin=0.14in]
    \item \textbf{BLEU, ROUGE-L, METEOR:} We utilize these n-gram or unigram matching metrics as primary diversity indicators. \vspace{-0.2cm}
    \item \textbf{Sent-sim:} To quantify semantic similarity, we first convert generated outputs into embedding vectors using Sentence-BERT, then compute the pairwise cosine similarity between these vectors. \vspace{-0.2cm}
    \item \textbf{LLM-Score:} We use OpenAI's recently revealed powerful reasoning model, GPT-o4-mini \cite{openai_gpt_o4_mini_2025}, to directly assess the diversity of multiple generated outputs for the same story prompt. See Appendix \ref{appendix:llmscore_eval} for more detailed rubric. \vspace{-0.2cm}
    \item \textbf{Degen:} We also use GPT-o4-mini to evaluate the level of degeneration in each generated output of the scale 0-1. The rubric for this evaluation consists of ``Syntactic Integrity (0.25)'', ``Semantic Trajectory (0.25)'', ``Lexical Sanity (0.25)'', and ``Noise Symptoms (0.25)''. See Appendix \ref{appendix:degen_eval} for the more detailed rubric.
\end{itemize}

\subsubsection{Quantitative Analysis of Diversity and Degeneration}
\label{main_result}
We apply our method and all baselines across three temperature values (low=0.7, moderate=1.0, high=1.3), four widely used  LLMs (Mistral 7B \cite{jiang2023mistral7b}, Llama 3B \cite{touvron2024llama3}, Llama 8B, Qwen 7B \cite{bai2023qwentechnicalreport}), and two story datasets that contain explicit story prompts (ReedsyPrompts, WritingPrompts). For each prompt, we generate 15 multi‑branch stories of 200 tokens each. We then evaluate diversity and degeneration metrics on these outputs. For reporting, we select the temperature setting that yields the greatest number of cases with an average degeneration score $\leq 0.1$ among the three temperature, since we empirically observe a sharp performance drop once the degeneration score exceeds $0.1$. If a model exceeds the average degeneration score 0.1 across all temperature settings, then we report the result of the lowest temperature setting. Due to space constraints, we only present the full results for Mistral-7B and the top-$8$ results for LLaMA-3B in this section. More comprehensive results can be found in Appendix \ref{appendix:comprehensive_results}.

As shown in Tables~\ref{main_rp_mistral7b}, \ref{main_wp_mistral7b}, \ref{main_rp_llama3b}, and \ref{main_wp_llama3b}, our model achieves the best performance on most metrics.  
Although Ours$_{CSP}$ exhibits higher diversity, it is not considered the best-performing model, as its degeneration rates exceed the acceptable 0.1 threshold across all temperature settings.
Ours$_{NSP}$, on the other hand, produces no degenerated samples but suffers from low overall diversity.
These results indicate that our Hybrid Penalty approach achieves an effective balance: It significantly enhances diversity compared to existing baselines while also demonstrating greater robustness to degeneration than the version relying solely on $CSP$.


Additionally, the significantly low diversity of CS and ACS may stem from modified instruction by accumulating previously generated stories in the input prompt as negative samples.
This leads language models to represent different hidden states for even identical tokens across stories. 
Consequently, even when the current candidate token is identical to one in the negative sample, their cosine similarity becomes small.
In contrast, our method uses the exact same instruction across branches, ensuring identical tokens yield identical hidden states and thus enabling accurate per-position penalties.

\subsubsection{Human Evaluation}
\begin{table}[t]
\centering
\small
\renewcommand{\arraystretch}{0.8}
\setlength{\tabcolsep}{8pt}
\begin{tabular}{lccc}
\toprule
method                          & \textbf{Diversity(↑)}            & \textbf{Creativity(↑)}           & \textbf{Degen(↑)}               \\
\midrule
Naive                   & 2.14 {\scriptsize± 0.8} & 2.33 {\scriptsize± 0.9} & 1.53 {\scriptsize± 1.3}   \\
Top-k                   & 3.38 {\scriptsize± 1.2} & 3.38 {\scriptsize± 1.5} & 3.94 {\scriptsize± 1.3} \\
Top-p                   & 2.71 {\scriptsize± 1.1} & 3.24 {\scriptsize± 1.5} & 4.00 {\scriptsize± 1.3} \\
Typical                 & 1.62 {\scriptsize± 0.7} & 2.81 {\scriptsize± 1.5} & 4.06 {\scriptsize± 1.0} \\
Mirostat                & 2.43 {\scriptsize± 1.1} & 3.14 {\scriptsize± 1.4} & \textbf{4.18} {\scriptsize± 1.2} \\
Min-p                   & 2.05 {\scriptsize± 0.9} & 3.10 {\scriptsize± 1.4} & 3.76 {\scriptsize± 1.3} \\
CS                      & 1.10 {\scriptsize± 0.5} & 2.95 {\scriptsize± 1.4} & 3.76 {\scriptsize± 1.2} \\
ACS                     & 1.14 {\scriptsize± 0.5} & 3.14 {\scriptsize± 1.4} & 3.71 {\scriptsize± 1.2} \\
DBS                     & 2.95 {\scriptsize± 1.2} & 3.29 {\scriptsize± 1.4} & 3.88 {\scriptsize± 1.4} \\
\midrule
GPT-4o                  & 2.62 {\scriptsize± 1.1} & 3.33 {\scriptsize± 1.5} & 4.00 {\scriptsize± 1.3} \\
\midrule
Ours$_{\text{CSP}}$     & 3.33 {\scriptsize± 1.4} & 3.19 {\scriptsize± 1.3} & 2.24 {\scriptsize± 1.2} \\
Ours$_{\text{NSP}}$     & 1.43 {\scriptsize± 0.6} & 2.62 {\scriptsize± 1.5} & 4.06 {\scriptsize± 1.0} \\
\midrule
\rowcolor{gray!15}
\textbf{Ours}           & \textbf{3.48} {\scriptsize± 1.3} & \textbf{3.71} {\scriptsize± 1.3} & 3.06 {\scriptsize± 1.4} \\
\bottomrule
\end{tabular}
\caption{Human evaluation scores across decoding methods (mean ± std). Best means in each column are in \textbf{bold}.}
\label{human_eval}
\end{table}

\paragraph{Human Evaluation.}
We recruit seven human annotators and let them rate each model’s outputs on a 5‑point Likert scale according to four criteria: \vspace{-0.2cm}
\begin{itemize}[leftmargin=0.14in]
  \item \textbf{Diversity:} The extent to which each story fundamentally differs from the others, beyond simple changes to character or theme names. \vspace{-0.2cm}
  \item \textbf{Degeneration:} The degree to which the text maintains grammatical correctness and lexical coherence without breakdown. \vspace{-0.2cm}
  \item \textbf{Creativity:} The overall originality and intrigue of the generated stories.
\end{itemize}
Annotators evaluate only the last five generated outputs (samples 11–15) from each set, which exhibit the most prominent diversity differences. For more detailed information about human evaluation, see Appendix \ref{appendix:human_eval}, \ref{correlation_appendix}.

As shown in Table \ref{human_eval}, our model achieves the highest scores from human annotators for both Diversity and Creativity, proving the effectiveness of the Similarity-based Contrastive Penalty in enhancing narrative richness. Furthermore, our model shows superior performance on Degeneration compared to Ours$_{CSP}$, demonstrating the enhanced robustness of the concept‑to‑narrative hybrid penalty against degeneration. These results align with the quantitative analysis in Section~\ref{main_result}, providing additional reliability.

\subsubsection{Hyper-parameter Search: T$_{0}$}

 We vary the inflection point hyperparameter T$_0$ (equation~\ref{gamma_cal}) from 5 to 100 to find the best value considering the trade‑off between diversity and text degeneration using Llama-3.1-8B. We fix the total generation length at 200 tokens. As shown in Figure \ref{fig:L0_test}, larger T$_0$ values tend to increase LLMScore and Degeneration score, with trendline slopes of 0.0012 and 0.0014, respectively. Considering this trade‑off,  we select an T$_0$ value of 25 as the optimal setting, as it yields the highest Diversity among outputs with Degeneration~$\leq$~0.1.

\subsubsection{Impact of Maximum Token}

 We observe how the degeneration scores of various models change as the maximum token increases using Llama-3.2-3B. As shown in Figure \ref{fig:max_token}, our model demonstrates greater robustness to degeneration than other sampling methods, only exceeding a degeneration score of 0.1 when the maximum token length reaches 2000, while other baselines exhibit a significantly higher degeneration score at shorter lengths.
These results indicate that, as sequence length grows, our method provides substantially higher robustness to degeneration than directly feeding accumulated outputs back into instructions.

\subsubsection{Analysis of Dormant Neuron Ratios}

To examine the range of neuron activation during iterative multi-branch story generation, we perform dormant neuron analysis~\cite{sokar2023dormant} using the LLaMA-3.1-8B model. Due to space limitations, the full set of results is presented in Appendix \ref{appendix:dormant_neuron}.  Specifically, we consider a neuron in a fully connected layer to be \textit{dormant} if its GELU activation falls below $5\times10^{-5}$. As shown in Figure \ref{fig:dormant_ratio}, all baseline methods exhibit progressively reduced neural activation, indicating a decline in latent creativity. However, our method exhibits a decreasing dormant neuron ratio, thus increasing the range of neuron activation as the number of iterations increases. These results suggest that it more effectively enhances the model’s inherent creativity, not by simply avoiding repetition of previously generated tokens.


\subsubsection{Qualitative Analysis}

Table \ref{tab:qualitative_analysis} illustrates how our method encourages conceptual and narrative divergence from a negative sample. In the example shown, the previously generated story (left) features a bright and cheerful setting (e.g., \textit{``beautiful summer evening''}, \textit{``bustling city''}) and expresses anticipation and nervousness. The new decoded story (right), by contrast, adopts a somber tone with a rainy setting (e.g., \textit{``drizzle''}, \textit{``nondescript bistro''}) and conveys emotions such as melancholy and boredom. The plot also shifts from a planned date to a coincidental bar encounter. These differences highlight the effectiveness of our method's Similarity-based Contrastive Penalty in promoting both conceptual- and narrative-diversity in multi-branch stories.

\section{Conclusion}

We introduce Avoidance Decoding, a novel decoding strategy designed to enhance the diversity of multi-branch story generation. Our method contrastively penalizes token logits based on similarity across different branch stories, using a hybrid of Concept-level and Narrative-level Similarity Penalties. Automatic and human evaluations consistently demonstrate that Avoidance Decoding significantly outperforms existing baselines in terms of diversity, while effectively mitigating the common trade-off between diversity and degeneration. Furthermore, dormant neuron analysis suggests that our method fosters deeper model creativity, as evidenced by broader neuron activation during generation.

%

\section{Limitations}
Although Avoidance Decoding demonstrates strong diversity and effective suppression of degeneration without any additional training or stochastic sampling, it has the drawback of increased decoding time. This issue becomes more pronounced as the number of negative samples increases. To mitigate the computational overhead, one potential solution is to store only a fixed-size window of recent outputs in the negative example memory rather than maintaining the entire output history. Furthermore, additional hyperparameter tuning may lead to better trade-offs between degeneration and diversity. 

\section{Acknowledgments}
We thank anonymous reviewers for their constructive and insightful comments. K. Jung is with ASRI, Seoul National University, Korea. This work was supported by the National Research Foundation of Korea(NRF) grant funded by the Korea goverment(MSIT) (RS-2025-02263628). This work was also partly supported by Institute of Information \& communications Technology Planning \& Evaluation (IITP) grant funded by the Korea government(MSIT) [RS-2021-II211343, Artificial Intelligence Graduate School Program (Seoul National University) \& No.RS-2021-II212068, Artificial Intelligence Innovation Hub \& No.RS-2022-II220184, Development and Study of AI Technologies to Inexpensively Conform to Evolving Policy on Ethics].

\bibliography{anthology,custom}
\bibliographystyle{acl_natbib}

\appendix

\section{Full Results of Dormant Neuron Analysis}
\label{appendix:dormant_neuron}

\begin{figure}[t]     
  \centering
  \includegraphics[width=\columnwidth]{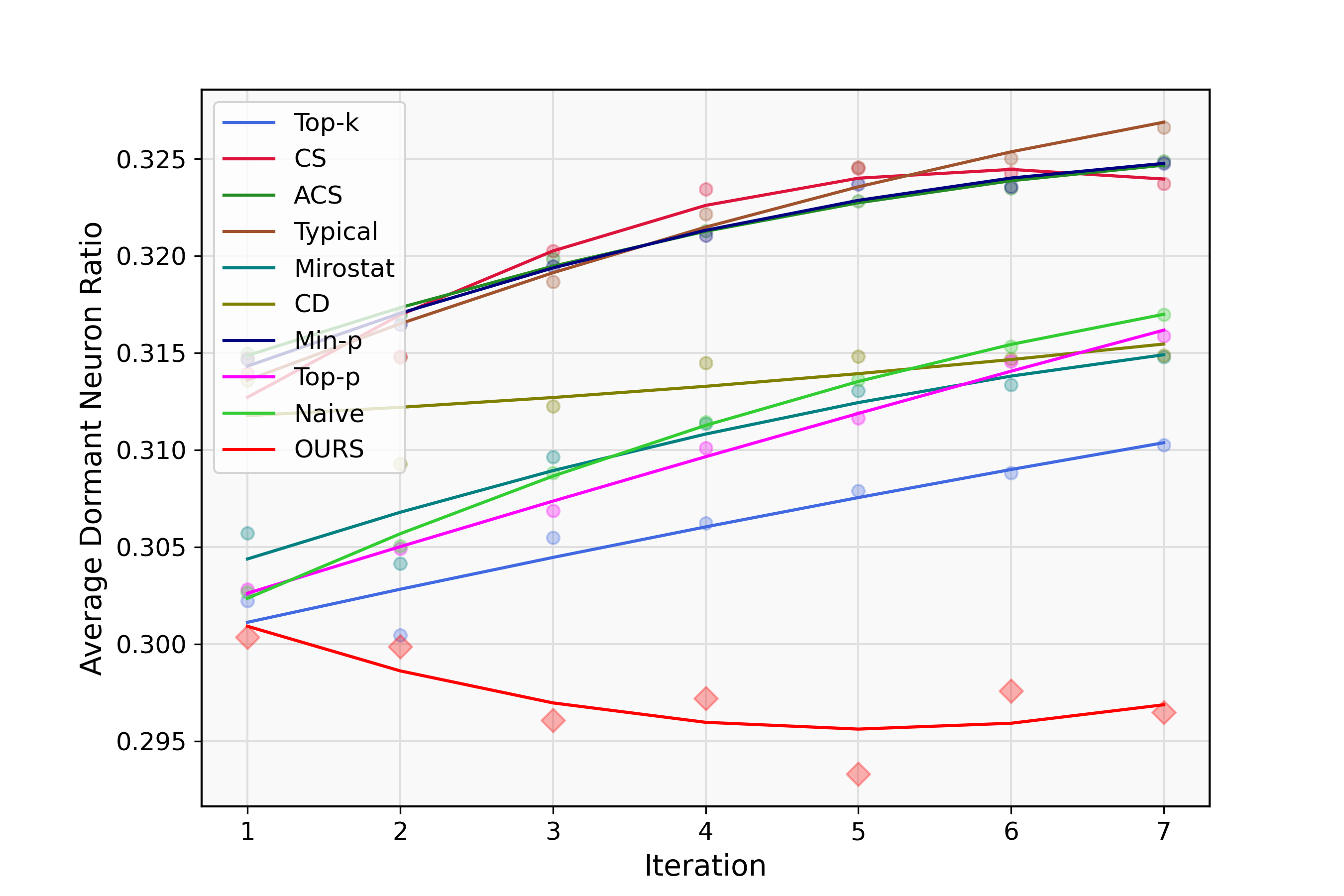}
  \caption{All dormant neuron ratios per iteration.}
  \label{fig:all_dormant_ratio}
\end{figure}

As shown in Fig.~\ref{fig:all_dormant_ratio}, among all baselines only our model exhibits a decrease in dormant neuron ratio over iterations. Specifically, the slopes of the linear trend lines are:
\[
\begin{aligned}
\text{Top-k}    &= +0.001540, \\ 
\text{CS}       &= +0.001875, \\ 
\text{ACS}      &= +0.001634, \\  
\text{Typical}  &= +0.002215, \\ 
\text{Mirostat} &= +0.001754, \\ 
\text{CD}       &= +0.000615, \\ 
\text{Min-p}    &= +0.001741, \\  
\text{Top-p}    &= +0.002260, \\ 
\text{Naive}    &= +0.002439, \\ 
\text{OURS}     &= -0.000675.
\end{aligned}
\]

\section{Degeneration Trend versus Iteration number}
\label{appendix:iteration_exp}

\begin{figure}[t]     
  \centering
  \includegraphics[width=\columnwidth]{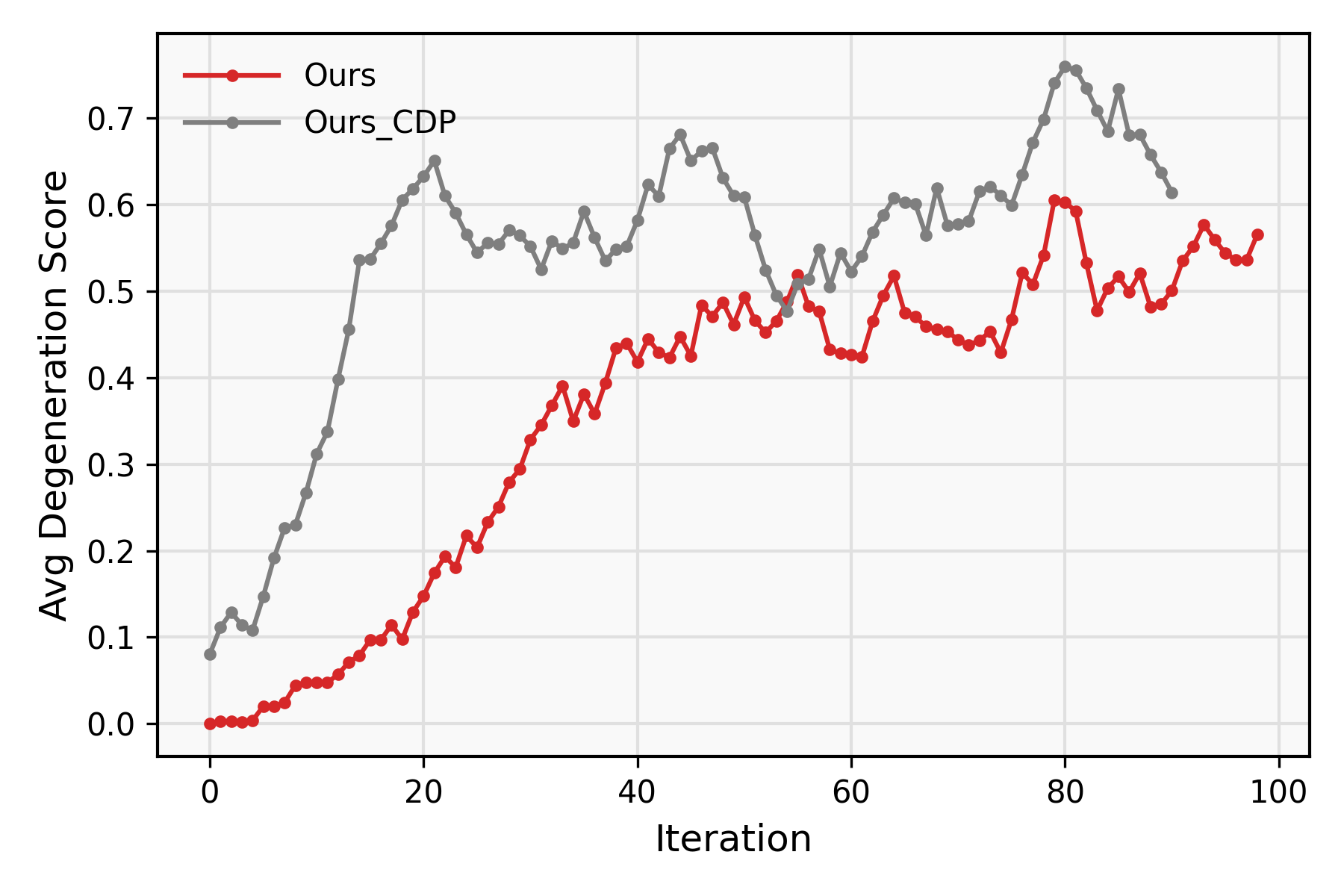}
  \caption{Smoothed average degeneration scores (window=10) versus iteration number.}
  \label{fig:iteration_plot}
\end{figure}

As shown in Figure \ref{fig:iteration_plot}, the average degeneration score increases as the number of iterations grows in large-branch settings. This trend suggests that the model’s linguistic structure gradually breaks down as it exhausts its intrinsic creative capacity due to the accumulation of previously generated outputs used as negative samples. Nevertheless, our model consistently yields lower degeneration scores than the ablated version using only CSP, demonstrating the robustness of the Hybrid Penalty even under high-branch configurations.

\section{Human Evaluation Details}
\label{appendix:human_eval}
We recruited graduate and undergraduate students fluent in English. The recruited annotators were provided with a detailed description of task definitions, instructions, and samples of each model. Also, all applicants were informed that their annotations would be used for academic purposes and would be published in paper material through the recruitment announcement and instructions. \\ Each of the seven annotators was given three samples—each consisting of five outputs from 13 baselines—and answered three questions for each sample.
 For the payment of the annotators, the co-authors conducted annotations for 5 hours first to estimate the average number of annotations that could be completed in the same time. Based on this estimation, a rate of 0.5 dollars per example was established to ensure that the annotators would be paid at least the minimum wage. 
 
To assess annotator agreement, we computed average‐measure ICC(2,k) values using each annotator’s model‐level mean scores: 0.93 for Diversity, 0.91 for Creativity, and 0.62 for Degen, indicating that all metrics achieved acceptable to excellent reliability.

\section{Comprehensive Experimental Results}
\label{appendix:comprehensive_results}
Table \ref{reedsyprompts_llama3b},  \ref{writingprompts_llama3b},
\ref{reedsyprompts_llama8b}, \ref{writingprompts_llama8b}, \ref{reedsyprompts_qwen7b},   and \ref{writingprompts_qwen7b} present the full experimental results on the ReedsyPrompts and WritingPrompts datasets using LLaMA 3B, LLaMA 8B, and Qwen 7B models. Our model achieves the best performance in most cases; however, on LLaMA 8B, its performance is highly competitive with that of Ours$_\text{CSP}$, with only a marginal difference.

\section{Degeneration Evaluation Details}
\label{appendix:degen_eval}

Figure \ref{fig:llmscore_degeneration_rubric} defines the rubric used to compute the \textbf{Degeneration} score, which measures the degree of degeneration in GPT-4o outputs. The evaluation incorporates four equally weighted dimensions: syntactic integrity, semantic coherence, lexical sanity, and noise symptoms. We specifically design the rubric to avoid false positives caused by poetic, metafictional, or stylistic sentences.

\section{LLMScore Evaluation Details}
\label{appendix:llmscore_eval}

Figure \ref{fig:llmscore_diversity_prompt} defines the prompt and rubric used to compute \textbf{LLMScore}, which evaluates the diversity of GPT-4o outputs.
 It emphasizes four key dimensions—\textit{perspective}, \textit{style}, \textit{plot structure}, and \textit{language variation}. We encourage generous recognition of creative or surface-level differences, even in the presence of shared semantic themes, ensuring fair reward for imaginative or structurally distinct outputs beyond token-level repetition.

\section{Detailed Instruction for Baseline Generation}
\label{appendix:instruction_baseline}

As shown in Figure~\ref{fig:conditional_prompt_negative}, we configure all baselines—except the Ours variants—to receive all prior generations along with an instruction that encourages dissimilar outputs in subsequent branches during repetitive multi-branch story generation.

\begin{table}[t]
  \centering
  \small
  \begin{tabular}{lcc}
    \hline
    \textbf{Metric} & \textbf{Spearman $r$} & \textbf{Pearson $r$} \\
    \hline
    BLEU ($\downarrow$)        & -0.73 & -0.68 \\
    ROUGE-L ($\downarrow$)     & -0.69 & -0.67 \\
    METEOR ($\downarrow$)      & -0.70 & -0.68 \\
    Cosine Sim ($\downarrow$)  & -0.72 & -0.72 \\
    LLMScore ($\uparrow$)      &  0.51 &  0.46 \\
    \hline
  \end{tabular}
  \caption{Correlation between metrics and human scores.}
  \label{tab:metric-correlations}
\end{table}


\begin{table}[t]
  \centering
  \small
  \begin{tabular}{cccc}
    \hline
    \textbf{$\beta$} & \textbf{$\delta$} & \textbf{Degen Score} & \textbf{LLMScore} \\
    \hline
    2.0 & 0.5 & 0.090 & 54.25 \\
    1.0 & 0.5 & 0.006 & 39.50 \\
    3.0 & 0.5 & 0.175 & 67.95 \\
    2.0 & 0.2 & 0.028 & 47.25 \\
    2.0 & 0.8 & 0.120 & 61.04 \\
    \hline
  \end{tabular}
  \caption{Degen Score and LLMScore across $(\beta, \delta)$ configurations.}
  \label{tab:beta-delta-results}
\end{table}

\begin{table*}[t]
\centering
\caption{ReedsyPrompts, Llama 3B}
\resizebox{\textwidth}{!}{
\begin{tabular}{lcccccc}
\toprule
\textbf{version}      & \textbf{BLEU(↓)} & \textbf{RougeL(↓)} & \textbf{METEOR(↓)} & \textbf{Sent-Sim(↓)} & \textbf{LLMScore(↑)} & \textbf{Degen} \\
\midrule
Naive                  &  8.40 & 21.22          & 27.60 & 52.53 & 30.50 & 0.00 \\
Top-k                  &  1.12 & \textbf{11.85} & 17.94 & 48.49 & 34.25 & 0.04 \\
Top-p                  &  6.22 & 17.96          & 24.34 & 52.50 & 29.35 & 0.00 \\
Typical                & 32.71 & 41.47          & 47.06 & 67.73 & 26.25 & 0.00 \\
Mirostat               &  9.88 & 22.48          & 28.90 & 55.71 & 24.90 & 0.04 \\
Min-p                  & 20.05 & 30.30          & 36.20 & 60.27 & 26.85 & 0.00 \\
CS                     & 66.75 & 70.64          & 73.33 & 80.60 & 21.10 & 0.00 \\
ACS                    & 59.78 & 64.91          & 67.68 & 77.95 & 19.00 & 0.00 \\
DBS                    & 10.39 & 21.30          & 30.20 & 61.83 & 26.25 & 0.01 \\
CD                     & 29.68 & 39.62          & 43.90 & 62.87 & 27.50 & 0.01 \\
\midrule
GPT-4o                 &  3.57 & 18.48          & 24.61 & 51.89 & 22.25 & 0.04 \\
\midrule
{Ours$_{CSP}$ }   &  0.85 & 11.61          & 14.06 & 31.73 & 60.10 & 0.23 \\
Ours$_{NSP}$        & 31.16 & 40.02          & 43.65 & 68.16 & 27.20 & 0.00 \\
\rowcolor{gray!20}
\textbf{Ours}          & \textbf{1.09} & 12.40 & \textbf{15.63} & \textbf{32.66} & \textbf{54.25} & 0.09 \\
\bottomrule
\end{tabular}}
\label{reedsyprompts_llama3b}
\end{table*}

\begin{table*}[t]
\centering
\caption{Writing Prompts, Llama 3B}
\resizebox{\textwidth}{!}{
\begin{tabular}{lcccccc}
\toprule
\textbf{version}                    & \textbf{BLEU(↓)} & \textbf{RougeL(↓)} & \textbf{METEOR(↓)} & \textbf{Sent-Sim(↓)} & \textbf{LLMScore(↑)} & \textbf{Degen} \\
\midrule
Naive                               & 10.45 & 22.32 & 29.59 & 60.01 & 29.50 & 0.00 \\
Top-k                               &  1.12 & 12.18 & 18.53 & 51.74 & 32.00 & 0.05 \\
Top-p                               &  4.66 & 16.98 & 23.67 & 55.70 & 27.00 & 0.08 \\
Typical                             & 41.65 & 47.65 & 53.21 & 79.54 & 19.00 & 0.01 \\
Mirostat                            & 29.35 & 37.53 & 43.50 & 71.27 & 21.00 & 0.00 \\
Min-p                               & 42.59 & 48.20 & 53.44 & 77.84 & 17.90 & 0.00 \\
CS                                  & 61.05 & 64.72 & 68.27 & 82.43 & 18.75 & 0.00 \\
ACS                                 & 68.96 & 72.09 & 73.45 & 86.12 & 18.50 & 0.01 \\
DBS                                 &  9.31 & 20.50 & 29.60 & 63.60 & 25.50 & 0.01 \\
CD                                  & 29.08 & 37.91 & 43.91 & 67.57 & 21.75 & 0.02 \\
\midrule
GPT-4o                              &  3.57 & 19.71 & 26.11 & 58.97 & 24.15 & 0.00 \\
\midrule
{Ours$_{CSP}$  }    &  0.82 & 11.54 & 14.03 & 32.96 & 57.25 & 0.13 \\
Ours$_{NSP}$          & 28.60 & 38.13 & 42.54 & 69.82 & 23.25 & 0.00 \\
\rowcolor{gray!20}
\textbf{Ours}                      & \textbf{1.05} & \textbf{11.68} & \textbf{15.10} & \textbf{34.03} & \textbf{50.60} & 0.09 \\
\bottomrule
\end{tabular}}
\label{writingprompts_llama3b}
\end{table*}

\begin{table*}[t]
\centering
\caption{ReedsyPrompts, Llama 8B}
\resizebox{\textwidth}{!}{
\begin{tabular}{lcccccc}
\toprule
\textbf{version}           & \textbf{BLEU(↓)} & \textbf{RougeL(↓)} & \textbf{METEOR(↓)} & \textbf{Sent-Sim(↓)} & \textbf{LLMScore(↑)} & \textbf{Degen} \\
\midrule
Naive                       &  8.96 & 21.71 & 28.73 & 53.61 & 22.95 & 0.00 \\
Top-k                       &  1.45 & 12.23 & 19.15 & 52.20 & 27.75 & 0.02 \\
Top-p                       &  4.49 & 16.90 & 24.05 & 51.97 & 28.75 & 0.00 \\
Typical                     & 29.83 & 39.26 & 45.72 & 63.11 & 21.10 & 0.00 \\
Mirostat                    &  8.88 & 21.89 & 28.45 & 54.65 & 28.25 & 0.00 \\
Min-p                       & 14.04 & 26.07 & 32.90 & 57.77 & 20.75 & 0.04 \\
CS                          & 49.13 & 56.86 & 60.38 & 67.30 & 24.85 & 0.00 \\
ACS                         & 50.81 & 58.29 & 61.63 & 68.98 & 20.25 & 0.01 \\
DBS                         &  9.66 & 21.15 & 29.67 & 62.08 & 29.90 & 0.01 \\
CD                          & 17.29 & 28.94 & 35.05 & 53.30 & 27.00 & 0.00 \\
\midrule
GPT-4o                      &  3.57 & 18.48 & 24.61 & 51.89 & 22.25 & 0.04 \\
\midrule
Ours$_{CSP}$           & \textbf{1.00} & \textbf{11.93} & \textbf{15.71} & \textbf{35.05} & 58.30 & 0.08 \\
Ours$_{NSP}$             & 32.14 & 41.04 & 45.59 & 70.56 & 24.15 & 0.00 \\
\rowcolor{gray!20}
\textbf{Ours}               &  1.19 & 12.53 & 16.94 & 35.73 & \textbf{58.60} & 0.10 \\
\bottomrule
\end{tabular}}
\label{reedsyprompts_llama8b}
\end{table*}

\begin{table*}[t]
\centering
\caption{Writing Prompts, Llama 8B}
\resizebox{\textwidth}{!}{
\begin{tabular}{lcccccc}
\toprule
\textbf{version}      & \textbf{BLEU(↓)} & \textbf{RougeL(↓)} & \textbf{METEOR(↓)} & \textbf{Sent-Sim(↓)} & \textbf{LLMScore(↑)} & \textbf{Degen} \\
\midrule
Naive                  & 11.88 & 23.82 & 30.55 & 63.89 & 25.35 & 0.00 \\
Top-k                  &  1.69 & 12.68 & 19.36 & 55.48 & 27.00 & 0.02 \\
Top-p                  & 21.76 & 32.18 & 38.47 & 68.25 & 22.50 & 0.00 \\
Typical                & 32.39 & 46.06 & 51.93 & 75.99 & 21.50 & 0.00 \\
Mirostat               & 14.64 & 25.71 & 33.28 & 65.24 & 23.75 & 0.00 \\
Min-p                  & 17.80 & 28.65 & 35.61 & 67.99 & 21.25 & 0.07 \\
CS                     & 64.48 & 68.59 & 71.30 & 78.95 & 17.75 & 0.00 \\
ACS                    & 65.76 & 68.93 & 72.48 & 82.56 & 20.25 & 0.00 \\
DBS                    & 10.21 & 21.60 & 30.35 & 65.40 & 23.80 & 0.03 \\
CD                     & 31.45 & 42.18 & 47.56 & 70.16 & 22.00 & 0.01 \\
\midrule
GPT-4o                 &  3.57 & 18.48 & 24.61 & 51.89 & 22.25 & 0.00 \\
\midrule
{Ours$_{CSP}$  } & \textbf{ 0.98} &\textbf{ 12.09} & \textbf{15.76} & 39.77 & \textbf{49.15} & 0.07 \\
Ours$_{NSP}$             & 35.58 & 43.69 & 48.02 & 75.59 & 20.65 & 0.00 \\
\rowcolor{gray!20}
\textbf{Ours}          &  1.25 & 12.15 & 16.50 & \textbf{38.05} & 49.00 & 0.07 \\
\bottomrule
\end{tabular}}
\label{writingprompts_llama8b}
\end{table*}

\begin{table*}[t]
\centering
\caption{ReedsyPrompts, Qwen 7B}
\resizebox{\textwidth}{!}{
\begin{tabular}{lcccccc}
\toprule
\textbf{version}      & \textbf{BLEU(↓)} & \textbf{RougeL(↓)} & \textbf{METEOR(↓)} & \textbf{Sent-Sim(↓)} & \textbf{LLMScore(↑)} & \textbf{Degen} \\
\midrule
Naive                  &  2.89 & 14.69 & 20.12 & 49.66 & 30.75 & 0.00 \\
Top-k                  &  2.00 & 13.75 & 19.88 & 51.91 & 27.75 & 0.00 \\
Top-p                  &  6.77 & 17.96 & 24.34 & 52.50 & 29.35 & 0.00 \\
Typical                &  3.64 & 41.47 & 47.06 & 52.52 & 25.75 & 0.03 \\
Mirostat               & 19.21 & 32.01 & 36.74 & 63.94 & 22.85 & 0.01 \\
Min-p                  & 14.54 & 28.40 & 33.26 & 61.97 & 25.90 & 0.01 \\
CS                     & 76.32 & 81.64 & 82.14 & 84.39 & 14.25 & 0.00 \\
ACS                    & 70.98 & 76.83 & 77.86 & 81.86 & 16.00 & 0.00 \\
DBS                    &  7.87 & 20.29 & 28.37 & 58.22 & 30.50 & 0.03 \\
\midrule
GPT-4o                 &  3.57 & 18.48 & 24.61 & 51.89 & 22.25 & 0.04 \\
\midrule
{Ours$_{CSP}$ }   &  0.70 & 10.07 & 10.67 & 30.37 & 63.75 & 0.19 \\
Ours$_{NSP}$        & 16.75 & 28.67 & 33.60 & 57.50 & 24.15 & 0.00 \\
\rowcolor{gray!20}
\textbf{Ours}          & \textbf{1.59} & \textbf{12.40} & \textbf{15.63} & \textbf{35.13} & \textbf{54.25} & 0.09 \\
\bottomrule
\end{tabular}}
\label{reedsyprompts_qwen7b}
\end{table*}

\begin{table*}[t]
\centering
\caption{Writing Prompts, Qwen 7B}
\resizebox{\textwidth}{!}{
\begin{tabular}{lcccccc}
\toprule
\textbf{version}                    & \textbf{BLEU(↓)} & \textbf{RougeL(↓)} & \textbf{METEOR(↓)} & \textbf{Sent-Sim(↓)} & \textbf{LLMScore(↑)} & \textbf{Degen} \\
\midrule
Naive                               &  6.08 & 18.34 & 23.78 & 56.14 & 24.75 & 0.01 \\
Top-k                               &  3.05 & 15.27 & 21.44 & 55.98 & 23.25 & 0.00 \\
Top-p                               & 18.44 & 30.91 & 35.70 & 65.52 & 24.25 & 0.00 \\
Typical                             &  5.51 & 17.12 & 23.21 & 55.74 & 26.00 & 0.04 \\
Mirostat                            & 26.32 & 38.59 & 43.11 & 71.10 & 20.50 & 0.01 \\
Min-p                               & 29.05 & 41.20 & 45.00 & 73.36 & 23.25 & 0.01 \\
CS                                  & 83.32 & 87.27 & 87.75 & 87.66 & 14.25 & 0.02 \\
ACS                                 & 77.46 & 82.98 & 83.03 & 84.68 & 17.25 & 0.00 \\
DBS                                 &  9.35 & 21.83 & 29.97 & 63.61 & 25.00 & 0.01 \\
CD                                  & 31.45 & 42.18 & 47.56 & 70.16 & 22.00 & 0.02 \\
\midrule
GPT-4o                              &  4.67 & 19.71 & 26.11 & 58.97 & 24.15 & 0.00 \\
\midrule
{Ours$_{CSP}$  }     &  1.30 & 12.20 & 13.37 & 36.18 & 48.00 & 0.13 \\
Ours$_{NSP}$          & 20.61 & 32.27 & 36.92 & 66.77 & 24.60 & 0.00 \\
\rowcolor{gray!20}
\textbf{Ours}                      & \textbf{1.63} & \textbf{12.08} & \textbf{14.93} & \textbf{37.40} & \textbf{49.25} & 0.06 \\
\bottomrule
\end{tabular}}
\label{writingprompts_qwen7b}
\end{table*}

\begin{figure*}[t]
\centering
\setlength{\fboxsep}{10pt}
\noindent\fbox{%
\parbox{\textwidth}{%
\textbf{You are a rigorous text degeneration evaluator.}

You will be given ONE passage of text. Your task is to assess whether it exhibits signs of LLM degeneration.

Return ONLY valid JSON with:
\begin{itemize}
  \item \texttt{"degeneration\_score"}: float in [0.0, 1.0]
  \item \texttt{"label"}: \texttt{"OK"} or \texttt{"DEGENERATED"} (threshold: 0.30)
  \item \texttt{"issues"}: a list of up to 4 bullet-point style strings, explaining key problems if any
\end{itemize}

\textbf{Use the following \underline{rubric}. Each of the 4 dimensions contributes equally (0.25 weight each) to the final score.}
\begin{itemize}
\item[--] \textbf{1. Syntactic Integrity (grammar \& sentence structure)} \\
\quad • 0.0 = flawless grammar, logical punctuation, complete sentences \\
\quad • 0.2–0.4 = minor run-ons or awkward phrasing \\
\quad • 0.6–0.8 = frequent structural errors or confusing syntax \\
\quad • 1.0 = broken or chaotic sentence structure \\
\quad \textit{Do NOT penalize complex or poetic sentence structures if grammatically valid.}

\vspace{1mm}
\item[--] \textbf{2. Semantic Trajectory (logical \& thematic coherence)} \\
\quad • 0.0 = clear, logical development of ideas \\
\quad • 0.2–0.4 = some loose transitions or unexpected inserts \\
\quad • 0.6–0.8 = idea flow breaks down across sentences \\
\quad • 1.0 = no coherent message or progression \\
\quad \textit{Do NOT penalize metafictional commentary or stylistic digressions if meaning is preserved.}

\vspace{1mm}
\item[--] \textbf{3. Lexical Sanity (word choices \& in-context appropriateness)} \\
\quad • 0.0 = all words match the tone and meaning \\
\quad • 0.2–0.4 = some unusual word choices but interpretable \\
\quad • 0.6–0.8 = strange phrasing, tone mismatches, rare word combinations \\
\quad • 1.0 = nonsensical or surreal word combinations (e.g. “fractional nut satisfaction”) \\
\quad \textit{Do NOT penalize poetic, archaic, or stylized language if used intentionally.}

\vspace{1mm}
\item[--] \textbf{4. Noise Symptoms (repetition, rambling, word salad)} \\
\quad • 0.0 = no unusual patterns \\
\quad • 0.2–0.4 = light stylistic repetition or verbosity \\
\quad • 0.6–0.8 = distracting repetition, filler, or randomness \\
\quad • 1.0 = clear signs of uncontrolled generation: token loops, hallucinations, nonsense \\
\quad \textit{Do NOT, Never penalize if the \textbf{final sentence} of the passage is truncated — this is because of the length limits.}
\end{itemize}

\textbf{Important:}
\begin{itemize}
\item Use professional judgment to apply these scores.
\item Focus on detecting \textit{true} degeneration (e.g., broken logic, nonsense, hallucinated text).
\item Never return anything except the JSON object. No extra explanation or comments.
\end{itemize}
}}
\caption{Prompt and Rubric for Rigorous Degeneration Evaluation}
\label{fig:llmscore_degeneration_rubric}
\end{figure*}

\begin{figure*}[t]
\centering
\setlength{\fboxsep}{10pt}
\noindent\fbox{%
\parbox{\textwidth}{%
\textbf{You are a text diversity evaluator.}

Below are 15 numbered text samples. Your task is to assess how diverse they are in terms of \textbf{perspective, style, plot structure, and language variation.}

Your output must be a JSON object with:
\begin{itemize}
  \item \texttt{"diversity\_score"}: a float between 0.0 and 1.0 (where 0 = all samples are nearly identical, and 1 = samples are maximally diverse)
  \item \texttt{"justification"}: a one-sentence explanation of your reasoning
\end{itemize}

\textbf{Scoring guidance:}
\begin{itemize}
\item 0.0: All samples are structurally and semantically almost identical.
\item 0.1–0.3: Slight variation in phrasing or detail, but mostly follow the same template.
\item 0.4–0.6: Notable variation in perspective, tone, setting, or content development.
\item 0.7–0.9: Substantial differences in narrative framing, imaginative detail, or genre shifts.
\item 1.0: Samples are maximally different in form, function, and voice.
\end{itemize}

Be generous when minor shifts in character, setting, or literary device occur. Do not penalize shared themes if surface features differ meaningfully.

\textbf{Return only a valid JSON object and nothing else.}
}}
\caption{Prompt and Evaluation Rubric for Measuring Textual Diversity}
\label{fig:llmscore_diversity_prompt}
\end{figure*}

\begin{figure*}[t]
\centering
\setlength{\fboxsep}{10pt}
\noindent\fbox{%
\parbox{\textwidth}{%
\textbf{You are a helpful and creative assistant that always responds in English and avoids undesired responses.}

Please write a story from the following prompt.

Do \textbf{NOT} generate responses that resemble the following examples: \\
\texttt{\{all\_previous\_outputs\}}
}}
\caption{Prompt for Conditional Story Generation with Explicit Negative Constraint}
\label{fig:conditional_prompt_negative}
\end{figure*}

\section{Use of ChatGPT and Compliance with OpenAI's Terms}

We utilized \textbf{OpenAI’s ChatGPT} for limited assistance in refining the writing and formatting of this paper. All substantive contributions, including the core methodology, experiments, and analysis, were conducted independently by the authors.

Our usage complies with \href{https://openai.com/policies/terms-of-use}{OpenAI’s Terms of Use} and \href{https://openai.com/policies/usage-policies}{Usage Policies}.

\section{Correlation Between Automatic Metrics and Human Ratings}
\label{correlation_appendix}
As shown in the Table \ref{tab:metric-correlations}, we conducted a correlation analysis between human ratings and automatic metrics. Since human evaluators assigned higher scores to outputs with greater diversity, in contrast to similarity-based metrics such as BLEU, ROUGE-L, METEOR, and Cosine Similarity, but in line with LLMScore, we conclude that each metric exhibits a strong correlation with human judgments.

Additionally, we did measure inter-annotator agreement among our human evaluators, which showed strong consistency, thereby supporting further reliability of our human evaluations.


\section{Empirical Experiments for Tuning Hyperparameters}
\label{beta_delta_appendix}
As shown in Table \ref{tab:beta-delta-results}, we conducted several empirical trials to determine hyperparameter values.
Although some configurations yield higher LLMScore, they also result in degeneration scores above 0.1, which we consider to indicate serious performance degradation.

\end{document}